\documentclass[10pt,twocolumn,letterpaper]{article}

\usepackage{cvpr}              
\usepackage{graphicx}
\usepackage{pdflscape}  

\definecolor{cvprblue}{rgb}{0.21,0.49,0.74}
\usepackage[pagebackref,breaklinks,colorlinks,allcolors=cvprblue]{hyperref}

\def\paperID{*****} 
\def\confName{CVPR}
\def\confYear{2026}

\title{Beyond Stationarity: Rethinking Codebook Collapse in Vector Quantization
}

\author{
Hao Lu$^{1}$\thanks{Corresponding author: Hao.Lu@advocatehealth.org} \quad
Onur C. Koyun$^{1}$ \quad
Yongxin Guo$^{1}$ \quad
Zhengjie Zhu$^{1}$ \quad
Abbas Alili$^{1}$ \quad
Metin Nafi Gurcan$^{1}$\\[4pt]
$^{1}$Wake Forest University School of Medicine, Winston-Salem, NC, USA\\[4pt]
{\tt\small \{Hao.Lu, Onur.Koyun, Abbas.Alili\}@advocatehealth.org} \\
{\tt\small \{Yongxin.Guo, Zhengjie.Zhu, Metin.Gurcan\}@wfusm.edu}
}

\begin{document}
\maketitle
\begin{abstract}
Vector Quantization (VQ) underpins many modern generative frameworks such as VQ-VAE, VQ-GAN, and latent diffusion models. Yet, it suffers from the persistent problem of codebook collapse, where a large fraction of code vectors remains unused during training. This work provides a new theoretical explanation by identifying the non-stationary nature of encoder updates as the fundamental cause of this phenomenon. We show that as the encoder drifts, unselected code vectors fail to receive updates and gradually become inactive. To address this, we propose two new methods: Non-Stationary Vector Quantization (NS-VQ), which propagates encoder drift to non-selected codes through a kernel-based rule, and Transformer-based Vector Quantization (TransVQ), which employs a lightweight mapping to adaptively transform the entire codebook while preserving convergence to the k-means solution. Experiments on the CelebA-HQ dataset demonstrate that both methods achieve near-complete codebook utilization and superior reconstruction quality compared to baseline VQ variants, providing a principled and scalable foundation for future VQ-based generative models. 
The code is available at: \url{https://github.com/CAIR-LAB-WFUSM/NS-VQ-TransVQ.git}

\end{abstract}

\section{Introduction}
\label{sec:intro}
\begin{figure}
  \centering
  \begin{subfigure}{1\linewidth}
    \centering
    \includegraphics[width=\linewidth]{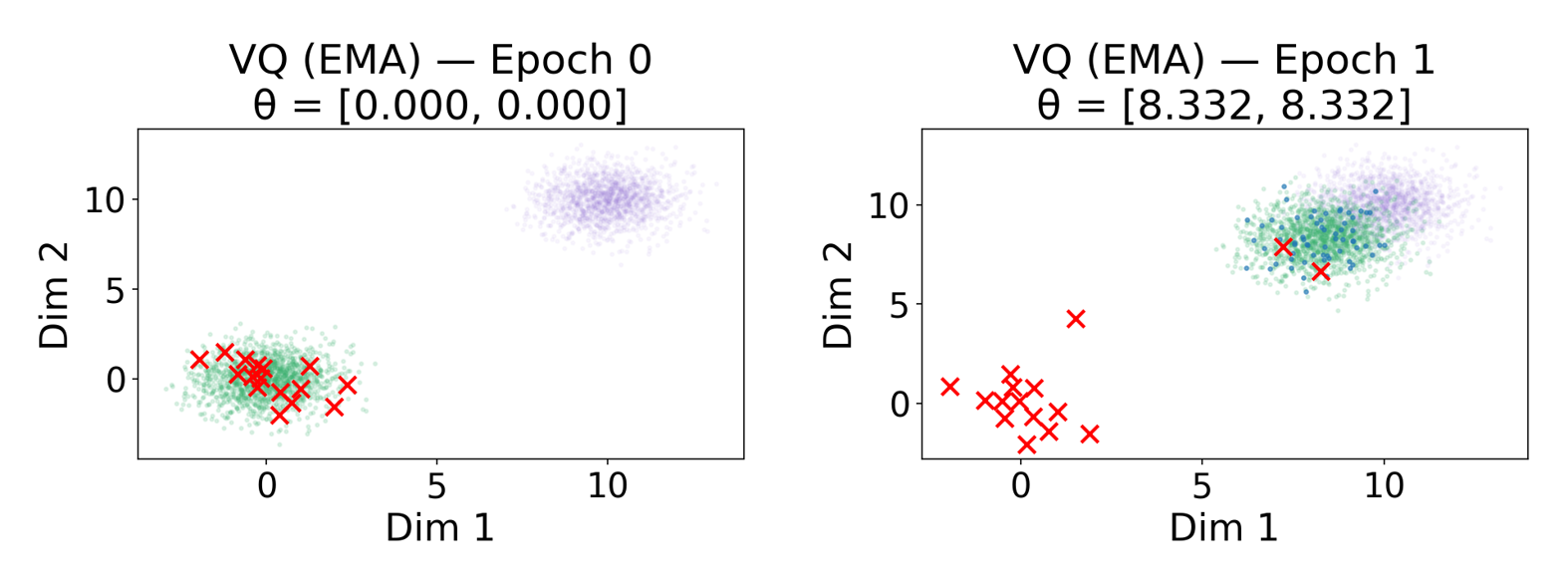}
    \caption{A toy example of vanilla VQ in non-stationary process.}
    \label{fig:vanila_vq_toy}
  \end{subfigure}
  \hfill
  \begin{subfigure}{1\linewidth}
    \centering
    \includegraphics[width=\linewidth]{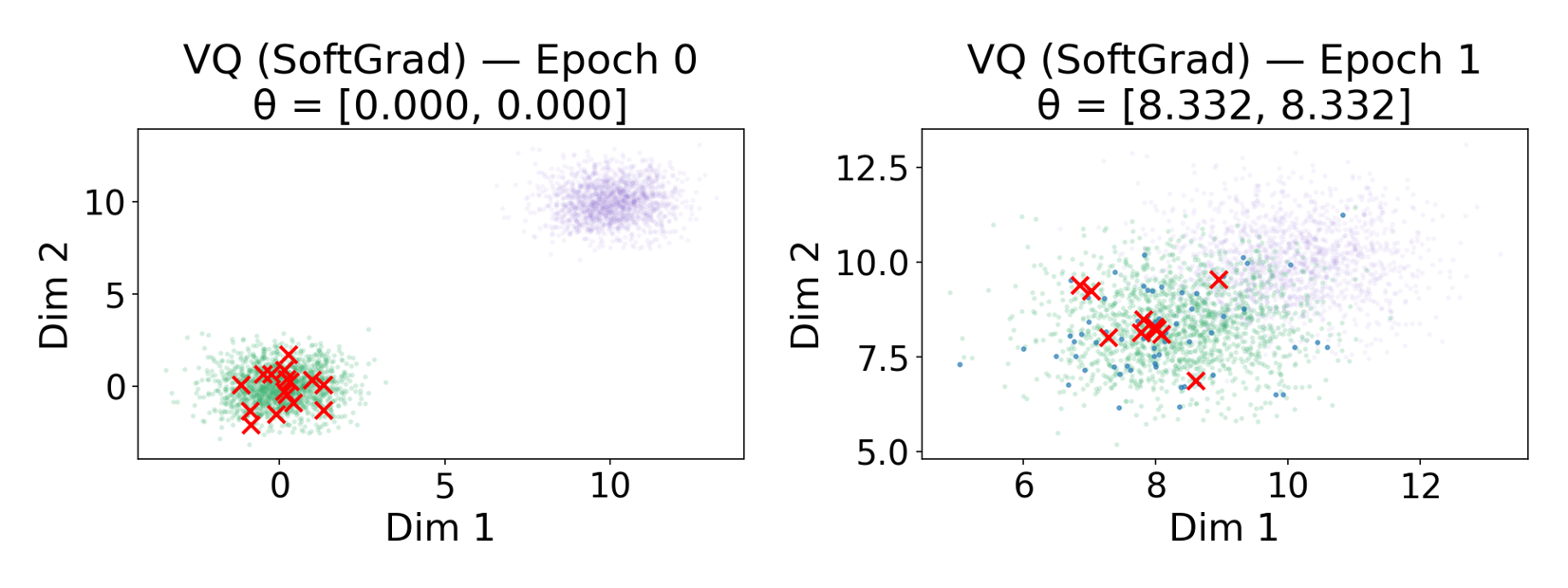}
    \caption{A toy example of proposed NS-VQ in non-stationary process.}
    \label{fig:nsvq_toy}
  \end{subfigure}
   \hfill
  \begin{subfigure}{1\linewidth}
    \centering
    \includegraphics[width=\linewidth]{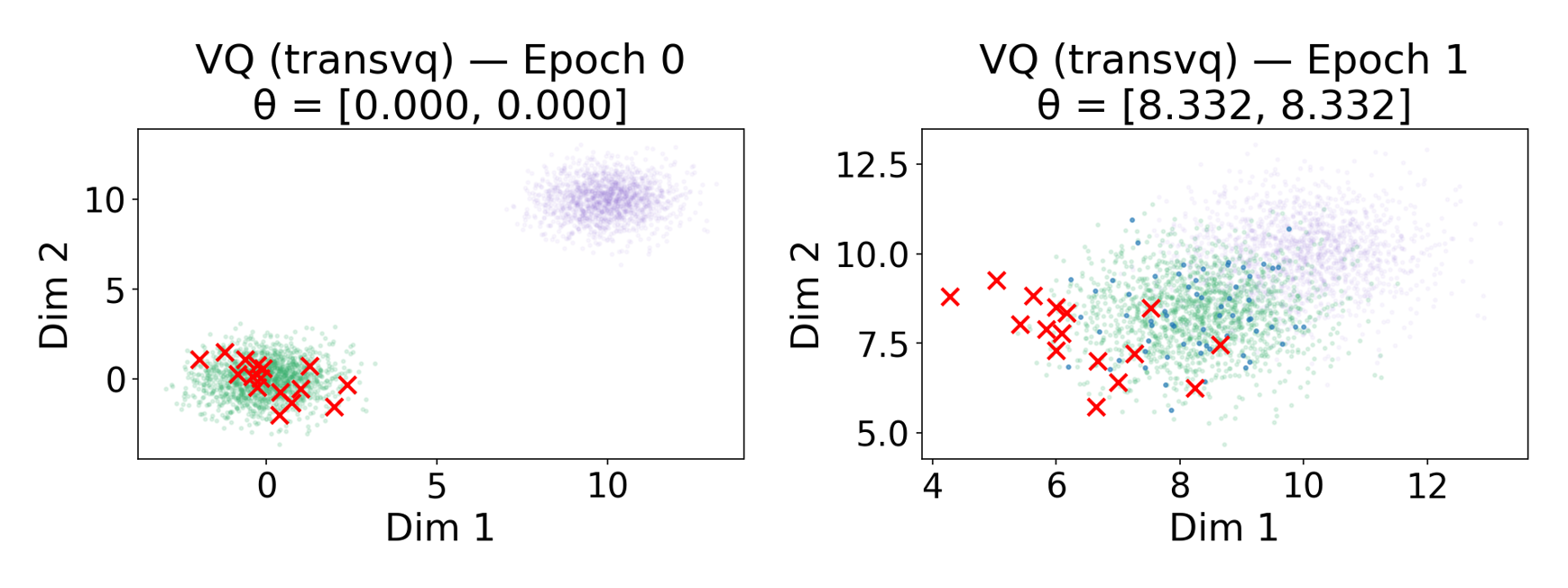}
    \caption{A toy example of proposed TransVQ in non-stationary process.}
    \label{fig:nsvq_toy}
  \end{subfigure}
  \caption{Illustration of codebook adaptation under non-stationary data. 
  (a) In vanilla VQ, codewords fail to track the drifting data distribution, leading to representation collapse. 
  (b) In the proposed NS-VQ, adaptive variance-controlled updates allow the codebook to follow distributional shifts over time, maintaining coverage and stability.
  (c) In the proposed TransVQ, a few codewords still lag behind the drifting data distribution, but the projector gradients drive all other codewords to move jointly toward the data, improving overall alignment.
  In the figures, purple dots denote the target distribution $Y_t$ (least visible), green dots represent the base data $X$, blue dots indicate the current batch $X_b$, and red crosses mark the codebook vectors $C$.
  }
  \label{fig:1}
\end{figure}

Vector Quantization (VQ) has long been a fundamental technique in signal processing, offering a way to represent continuous data by discrete codes \cite{RN1}. Since its successful integration into Variational Autoencoders (VQ-VAE) by \cite{RN2}, VQ has become the backbone of many state-of-the-art generative frameworks, including VQ-GAN \cite{RN3} and Latent Diffusion Models (LDM) \cite{RN4}. In addition, VQ-based tokenization is widely adopted in visual-language models (VLMs) \cite{RN5}, where images are discretized into code sequences that can be processed by large language models.

Despite these successes, VQ models face a persistent and widely recognized challenge: codebook collapse. As the codebook size increases, a large portion of code vectors remain inactive during training, leading to poor code utilization. This severely limits the effectiveness of VQ in large-scale generative modeling, where scaling the codebook is expected to improve representation power. While numerous practical solutions have been proposed—such as stochastic quantization, codebook reset strategies, distribution regularization, or external feature initialization—most of them are heuristic in nature. They improve utilization rates in practice but lack theoretical justification. Consequently, even when utilization approaches 100\%, the final performance across methods varies significantly. This indicates that our current understanding of the codebook collapse problem remains largely empirical and intuition-driven.

In this paper, we take a new theoretical perspective by analyzing the non-stationary nature of VQ-VAE. We show that encoder updates inherently make the latent representation a non-stationary process, which explains why codebook entries may fail to receive updates and eventually collapse, (see Fig.~\ref{fig:1}). This connection between non-stationarity and codebook collapse provides, for the first time, a theoretical foundation for understanding the phenomenon. 

\begin{figure*}
  \centering
  \begin{subfigure}{0.8\linewidth}
    \centering
    \includegraphics[width=\linewidth]{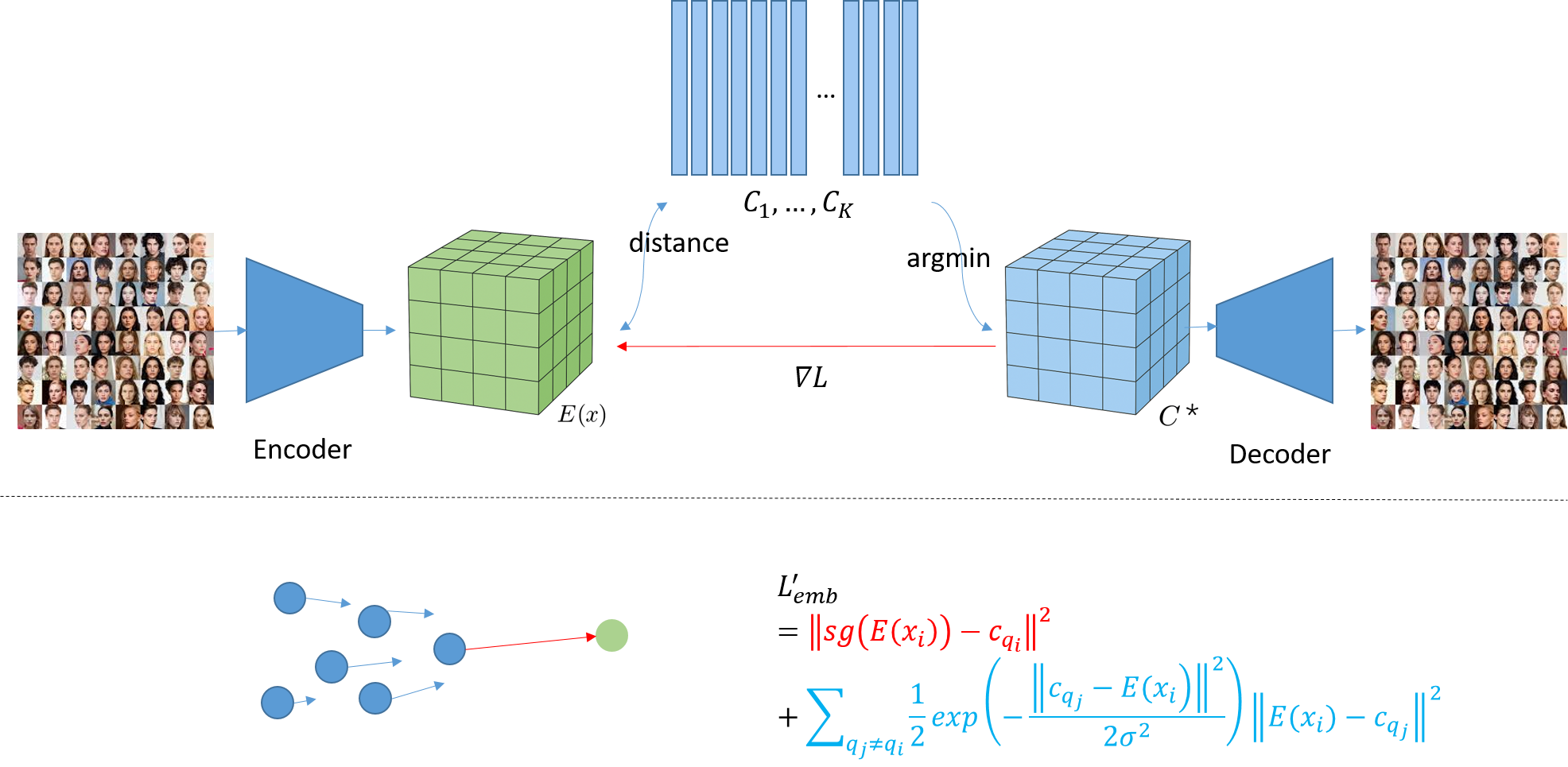}
    \caption{Scheme of the proposed NS-VQ module, which models non-stationary codebook dynamics through variance-controlled updates.}
    \label{fig:2a}
  \end{subfigure}
  \hfill
  \begin{subfigure}{0.8\linewidth}
    \centering
    \includegraphics[width=\linewidth]{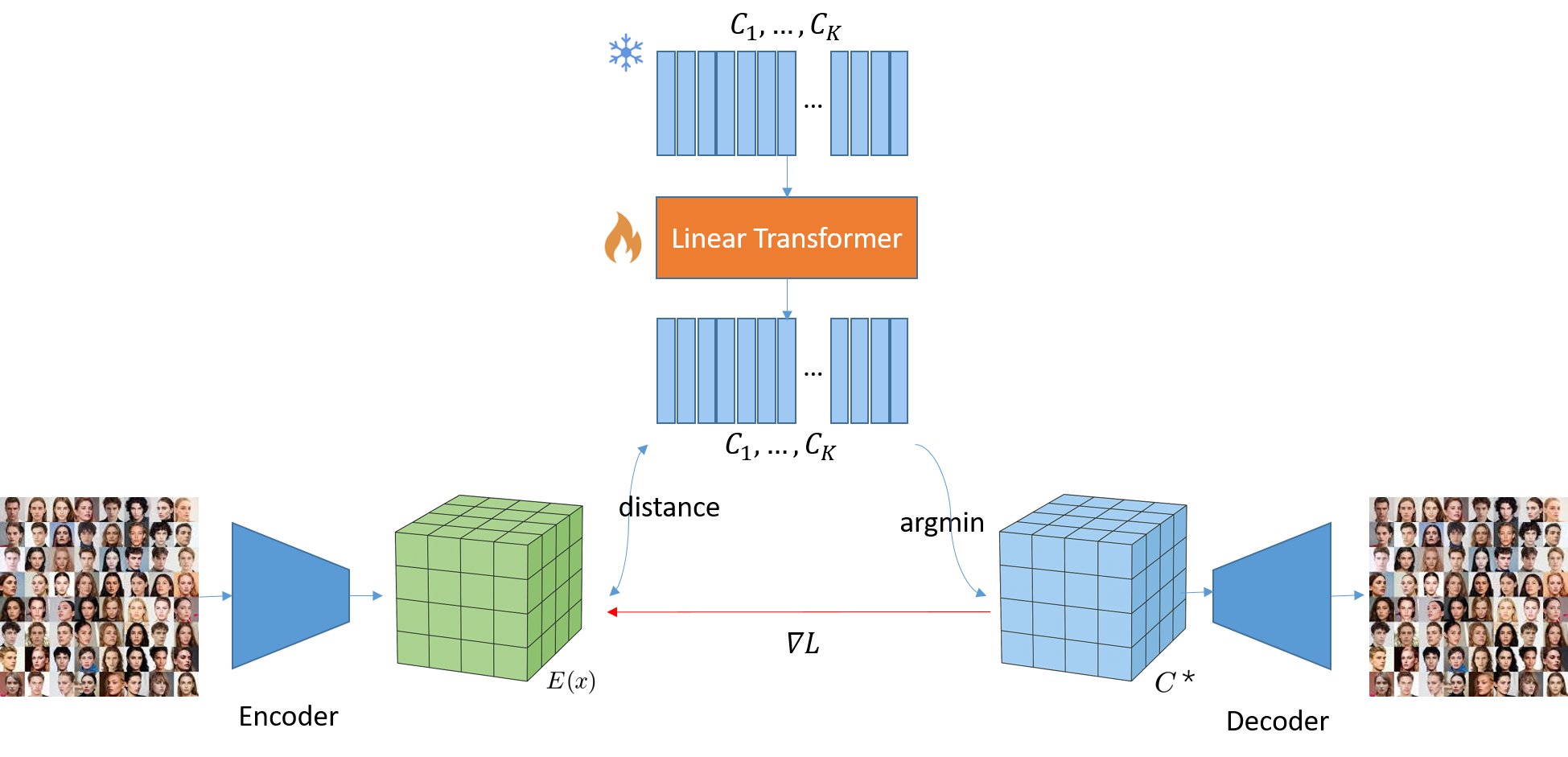}
    \caption{Scheme of TransVQ, which introduces transformer-based codebook interaction for non-stationary quantization.}
    \label{fig:transvq}
  \end{subfigure}
  \caption{Scheme of the proposed (a) Non-Stationary Vector Quantization (NS-VQ) and (b) the Transformer-based Vector Quantization (TransVQ). 
  }
  \label{fig:2}
\end{figure*}

Building on this insight, we propose two new VQ methods designed to address codebook collapse:
\begin{itemize}
    \item Non-Stationary VQ (NS-VQ), illustrated in Fig.~\ref{fig:2a}: introduces a kernel-based approximation that propagates encoder drift to non-selected codes, thereby increasing codebook utilization without breaking theoretical convergence conditions.
    \item Transformer-based VQ (TransVQ), illustrated in Fig.~\ref{fig:transvq}: employs a lightweight mapping function $P_\varphi(\cdot)$ that adaptively transforms the entire codebook in response to encoder updates.
\end{itemize}

Unlike prior methods such as SimVQ that rely solely on linear mappings and lose convergence guarantees, TransVQ preserves the theoretical conditions for convergence to the k-means solution while ensuring smooth adaptation of all codes.
 
We evaluate these methods on image reconstruction tasks within the VQ-VAE framework. Experimental results show that both NS-VQ and TransVQ significantly improve reconstruction quality, as measured by rFID, LPIPS, and SSIM, while maintaining nearly complete codebook utilization. Furthermore, our analysis of batch-size effects empirically validates the theoretical predictions of our non-stationary framework.

Our contributions can be summarized as follows:
\begin{itemize}
    \item We provide a theoretical analysis showing that the non-stationary nature of encoder updates is the root cause of codebook collapse in VQ-VAE.
    \item We propose NS-VQ, a novel variant that propagates encoder drift to non-selected codes via a kernel-based update rule, ensuring improved utilization.
    \item We propose TransVQ, which leverages a learnable codebook mapping function that preserves k-means convergence conditions and effectively adapts the entire codebook.
    \item We conduct extensive experiments on image reconstruction benchmarks, demonstrating that both methods achieve higher utilization and superior reconstruction quality compared to baseline VQ methods.
\end{itemize}

\section{Related Work}
\label{sec:2}
Since its introduction as a stochastic approximation (SA) of k-means \cite{RN1} and later application to Variational Autoencoders (VQ-VAE) by \cite{RN2}, Vector Quantization (VQ) has become a fundamental building block in modern visual generative models and image tokenization frameworks. It serves as the core component in models such as VQ-GAN~\cite{RN6}, LDM~\cite{RN4}, and VLM~\cite{RN7}, bridging continuous visual features and discrete token representations ~\cite{RN8}. Alongside these applications, a rich body of research has explored improving the optimization and utilization of VQ in neural networks.
\subsection{Optimization strategies}
Several studies have sought to improve the optimization process of VQ models. Huh~\cite{RN9} proposed an alternating optimization scheme that separately updates the encoder and quantization layers to accelerate convergence, albeit with increased computational cost. Baevski et al. (2019)~\cite{RN10} introduced the Gumbel-Softmax trick, providing a differentiable approximation to the discrete argmin operation during codeword selection. More recently, the rotation trick~\cite{RN11} reformulated the mapping from latent features to codebook vectors as a joint rotation-and-scaling transformation, allowing gradient computation through geometric reparameterization.
\subsection{Improving codebook utilization}
Several approaches have been developed to address codebook collapse. Yu~\cite{RN3} proposed maintaining the codebook in a lower-dimensional latent space to increase utilization. SoundStream~\cite{RN12} introduced a dynamic replacement policy that substitutes rarely used codes with randomly sampled vectors from the current batch. SimVQ~\cite{RN13,RN14} reparametrized the codebook as a linear transformation CW, enabling gradients to flow to non-winner code vectors and achieving high utilization with a simple design. Other works improve utilization through additional constraints, such as orthogonality regularization~\cite{RN15} and cosine-similarity penalties~\cite{RN3}. In contrast, some methods move away from the stochastic k-means formulation entirely—for instance, FSQ~\cite{RN16}, which performs scale-based quantization, and LFQ~\cite{RN5}, which employs directional (sign-based) clustering.
\subsection{Alternative quantization approaches}
Beyond codebook optimization, extensions such as residual VQ~\cite{RN12} aim to reduce quantization error by applying multi-stage residual encoding.
Our work differs in focus: rather than introducing heuristic improvements, we theoretically analyze the cause of codebook collapse and propose two methods—NS-VQ and TransVQ—that preserve the convergence property of VQ to the k-means solution. SimVQ inspires TransVQ, yet we extend it with a theoretical explanation of why codebook mapping mitigates collapse and modify it to maintain k-means convergence. Finally, although our experiments employ the standard straight-through estimator (STE), both proposed methods are fully compatible with the Gumbel-Softmax and rotation-trick frameworks.

\section{Method}
\label{sec:3}
\subsection{Preliminaries}
A Vector-Quantized Variational Autoencoder (VQ-VAE) is a reconstructive encoder–decoder architecture that introduces a vector quantization (VQ) layer to convert continuous latent representations into discrete codes.
We assume that data $x \in R^d$ is sampled from dataset
$X=\{x_1,…,x_N\}$,
Let set the encoder of the VQ-VAE is denoted to
$E_\theta (x)\in \mathbb{R}^d$,
and
$C=(c_1,…,c_K )\in(R^d )^K$
where K is the number of code vectors.
We define the Jacobian matrix (linear approximation of encoder changes with respect to parameters) as
$J_\theta (x)=\frac{\partial E_\theta (x)}{\partial\theta}$
The Voronoi partition associated with codebook C is
$V_k (C)= \{x: ||x-c_k || \leq ||x-c_j ||,\forall j\}  $
For each sample $x_i$, its assigned code index is denoted by $q_i$, where $x_i\in V_{q_i}(C)$.

Since the quantization step is non-differentiable, the straight-through estimator (STE)~\cite{RN3} is applied to enable gradient flow. Specifically, the quantized representation is expressed as:
$$ Q_e (x_i )=E_\theta (x_i )+sg(c_{q_i}-E_\theta (x_i ))$$
where $Q_e$ is VQ layer output. This yields $(\partial Q_e (x_i))/\partial E(x_i ) =1. $

The overall training objective of VQ-VAE combines reconstruction and commitment losses:
\[ L = \log p(x \mid Q_e) + \| \operatorname{sg}(E_\theta(x)) - c_q \|_2^2 + \beta \| E_\theta(x) - c_q \|_2^2 \],
where the first term is typically the mean squared error (MSE):
\[
\log p(x \mid Q_e) = \| x - g_\psi(Q_e) \|_2^2
\]
where $g_\psi (\cdot)$ is the decoder of VAE.
Here, the commitment loss encourages the encoder to stay close to the chosen embedding, while the codebook loss ensures that code vectors move toward the encoder outputs.

\subsection{Non-stationarity of VQ-VAE and Its Effect on the Quantization Process}
We assume k-means initialization, i.e., at iteration $t=0$, which means initially $C^{(0)}$ fits
$c_k^{(0)}  = \mathbb{E}[x \mid x\in V_k(C) ]$, where $\mathcal{E}$ means expectation or mean,
so that $C^{(0)}$ is k-means solution of the latent space and it is also the locally optimal (a fixed point) of VQ algorithm (proof is in Supplementary 8
) if the encoder is frozen.
\subsubsection{Non-stationarity in Training}
In practice, the encoder parameters $\theta$ are updated during training, making $E_{\theta^{(t)}}(x)$  a time-varying, non-stationary random process. This non-stationarity is a key factor in the emergence of “dead” codebook entries in traditional VQ-VAE training.
For simplicity, assume batch size = 1 (the reasoning extends naturally to larger batch sizes).
When the first sample $x_1$ is processed, the encoder parameters are updated by backpropagation. Under the straight-through estimator (STE),
\begin{align}
   \Delta \theta ^{(0)}=\frac{\partial L}{\partial c_{q_1}^{(0)}} I \frac{\partial E_\theta (x_1)^{(0)}}{\partial \theta ^{(0)}} 
\end{align}

where $L$ is the reconstruction loss.
At the same time, the selected codebook entry is updated by the embedding loss:
\begin{align}
    L_\text{emb}^{(t)}=\frac{1}{d}||\operatorname{sg}(E_\theta (x_n)^{(t)})-c_{q_n}^{(t)}||^2 
\end{align}
leading to
\begin{align}
    c_{q_m}^{(1)} = 
    \begin{cases}
    \displaystyle
        c_{q_m}^{(0)}  + \eta_0  (x_1- c_{q_m}^{(0)}), & m = 1,  \\
        c_{q_m}^{(0)}, & m\neq 1
    \end{cases}
\end{align}

Consider the next sample $x_2$, assumed (without loss of generality) to belong to a different Voronoi cell than $x_1$. After the encoder update,
\begin{align}
    E_{\theta^{(1)}}(x_2)\approx E_{\theta ^{(0)}} (x_2) + J_{\theta ^{(0)}}(x_2)\Delta \theta^{(0)}
\end{align}
Since $c_{q_2}^{(1)}$   was not updated in the previous step,
$c_{q_2}^{(1)}=c_{q_2}^{(1)}$   
However, the Jacobian term may cause $E_{\theta ^{(1)}}(x_2)$ to fall outside its original Voronoi region $V_{q_2}(C)$. As a result, $c_{q_2}^{(1)}$ may never be selected for updates, eventually becoming a “dead” code.

This mechanism explains the early codebook collapse phenomenon observed in\cite{RN9} [Fig. 9]. We further illustrate this with toy examples (in supplementary 11
and 7
), showing how encoder shifts (translation, scaling, splitting) can cause persistent dead codes. Intuitively, as the encoder parameters evolve, the mapping from inputs to latent space shifts, causing previously used codebook entries to become obsolete.
\subsubsection{Effect of Batch Size}
From this reasoning, a larger batch size should alleviate codebook collapse: in the first few iterations, more code vectors receive updates, increasing codebook utilization. In the extreme case, if batch size equals dataset size, all code vectors would be updated each iteration, eliminating dead codes entirely.
We validated this intuition experimentally by testing VQ-VAE reconstruction performance under different batch sizes (see Fig.~\ref{fig:rfid_batchsize}).
\subsection{Method 1: Non-Stationary Vector Quantization (NS-VQ)}
We propose Non-Stationary Vector Quantization (NS-VQ) to address the problem of dead codes caused by encoder non-stationarity. NS-VQ modifies the codebook update rule so that unused codes also receive meaningful updates, thereby mitigating early collapse.
\subsubsection{Motivation}
At $t$ step the encoder is updated with input $x_i$, the representation of other samples like $x_j$ becomes
\begin{align}
    E_{\theta ^{(t)}}(x_j)=E_{\theta ^{(t)}}(x_j) + \Delta E_\theta(x_j) 
\end{align}
where we use a linear approximation to estimate the additional term
\begin{align}
    \Delta E_\theta(x_j)\approx J_{\theta ^{(t)}}(x_j)\Delta \theta^{(0)}
\end{align}
may cause $E_{\theta^{(t)}}(x_j)$ to fall outside its original Voronoi region, leaving its codebook entry $c_{q_j}$ unused.
To counter this, we add a corresponding update to the codebook:
\begin{align}
\Delta c_{q_j} =
\begin{cases}
\displaystyle \frac{\partial L_{\text{emb}}}{\partial c_{q_j}}, & q_j = q_i, \\[8pt]
\Delta E(x_j), & q_j \ne q_i.
\end{cases} 
\end{align}
where $L_{emb}$ is the standard embedding loss.
\subsubsection{Estimating $\Delta E(x)$}
Since only $x_i$ is available in the current batch, we approximate $\Delta E(x_j)$ for other codes $q_j$ using the gradient update from $x_i$.
From one gradient descent step:
\begin{align}
    \Delta \theta=-\eta \frac{\partial L(E_\theta (x_i))}{\partial \theta}=-\eta J_\theta (x_i )^T g_i
\end{align}
where
$(g_i=\frac{\partial L}{\partial E }/(E_\theta(x_i))$
Then, for any $x_j$:
\begin{align}
    \Delta E_\theta (x_j) &\approx J_\theta(x_j) \Delta \theta\\
    &=-\eta J_\theta(x_j) J_\theta (x_i)^T g_i
    \\
    &=-\eta K_\theta (x_j,x_i ) g_i
\end{align}
where $K_\theta (x_j,x_i )=J_\theta (x_j ) J_\theta (x_i )^T$
being the neural tangent kernel (NTK).
Using $\Delta \theta \approx J_\theta(x_i)^{-1} \Delta E_\theta (x_i)$ to approximate $\Delta \theta$, we can get
\begin{align}
\Delta E(x_j) \approx K_\theta(x_j, x_i)\, K_\theta(x_i, x_i)^{-1} \, \Delta E(x_i)
\end{align}
Since computing the NTK is intractable, we approximate it by a Gaussian RBF kernel:
\begin{align}
k(x_j, x_i) = \exp\!\left(-\frac{\|E(x_j) - E(x_i)\|^2}{2\sigma^2}\right)
\end{align}
Thus,
\begin{align}
    \Delta E(x_j) = \exp\!\left(-\frac{\|E(x_j) - E(x_i)\|^2}{2\sigma^2}\right)\Delta E(x_i)
\end{align}
In practice, we approximate the distance by replacing $E(x_j)$ with its code vector $c_{q_j}$, and Interestingly, we found that simple substitution $\Delta E(x_i)$ with $(x_i-c_{q_j})$ works better than the full NTK-based estimate. 
\begin{align}
    \Delta E(x_j)\approx \exp\!\left(-\frac{\|E(x_i)-c_{q_j} \|^2}{2\sigma^2}\right) (E(x_i)-c_{q_j})
\end{align}
This leads to a practical and effective update rule:
\begin{align}
    \Delta c_q =
    \begin{cases}
    \displaystyle 
    \frac{\partial L_{\text{emb}}}{\partial c_{q_j}}, & q_j = q_i, \\[8pt]
    \exp\!\left(-\frac{\|E(x_i)-c_{q_j} \|^2}{2\sigma^2}\right) (E(x_i)-c_{q_j}), & q_j \ne q_i.
    \end{cases} 
\end{align}
This update can be derived from an auxiliary embedding loss:
\begin{equation}
\begin{split}
L_{emb}^\prime
&= \| \operatorname{sg}(E(x_i)) - c_{q_i} \|^2 \\
&\quad + \sum_{q_j \neq q_i}
\exp\!\left(-\frac{\|E(x_i)-c_{q_j} \|^2}{2\sigma^2}\right)
(E(x_i) - c_{q_j})
\end{split}
\end{equation}
where $sg(\cdot)$ is the stop-gradient operator.
\subsubsection{Modified STE}
It is worth noting that we also revisit the update rule in \cite{RN9} [Eq. (21)], which proposed a “delay-free” correction term for updating $c_{q_i}^{(0)}$ based on $E(x_i)^{(1)}-c_{q_i}^{(0)}$. That method, however, required tuning an additional hyperparameter $v$, which limits its practical applicability.
In our approach, we re-derive a similar correction without the need for extra hyperparameters. The full derivation and proof are provided in the supplementary 9
Shortly, we updated straight-through estimator (STE) for better update $c_{q_i}$ as:
\begin{align}
    Q_e=E(x_i )+sg(c_{q_i}-E(x_i ))+\frac{2}{d} c_{q_i}-sg(\frac{2}{d}c_{q_i})
\end{align}
where $d$ is the multiplication of batch size and patch numbers at latent feature.
In summary, compared with traditional VQ, NS-VQ differs in three aspects:
\begin{itemize}
    \item New embedding loss introduces cross-updates for non-selected codes.
    \item Modified STE improves stability under encoder drift.
\end{itemize}
As shown in Fig.~\ref{fig:rfid_usage}, these changes significantly improve reconstruction quality, with further details provided in the next section.
\subsection{Method 2: Transformer-based Vector Quantization (TransVQ)}
We propose a second approach, called Transformer-based Vector Quantization (TransVQ). The idea is simple: if encoder updates add a drift term to feature representations, why not apply a learnable transformation to the codebook as well?
\subsubsection{Motivation}
For any sample $x_j$, the encoder update at iteration t can be written as
\begin{align}
    E_{\theta^{(t+1)}} (x_j )=E_{\theta^{(t)}} (x_j )+J_{\theta^{(t)}} (x_j )\Delta \theta^{(t)}
\end{align}
where
\begin{align}
    \Delta E_{\theta^{(t)}} (x_j) = J_{\theta^{(t)}}(x_j)\Delta \theta^{(t)}
\end{align}
Instead of leaving the codebook fixed, we introduce a mapping function $P_\phi(\cdot)$ that transforms the codebook:
\begin{align}
    C^\prime=P_\phi(C)
\end{align}

During training, only $\phi$ is updated, while the base codebook C remains unchanged. After an update on a batch $x_i$, not only is its corresponding transformed code $P_\phi(c_{q_i})$ updated, but all codes are simultaneously updated as
\begin{align}
    P_{\phi ^{(t+1)}} (c_{q_j} )=P_{\phi ^{(t)}} (c_{q_j} )+J_{\phi ^{(t)}} (c_{q_j} )\Delta \phi^{(t)}
\end{align}
where,
\begin{align}
    J_\phi(t)(c_{q_j})=\frac{\partial P_\phi^{(t)}(c_{q_j})}{\partial \phi ^{(t)}}
\end{align}

\subsubsection{Key Observation}
In principle, one could explicitly enforce the alignment between encoder and codebook updates by introducing an additional constraint term
\begin{align}
L_{align} = ||J_{\phi(t)}(c_{q_m})\Delta\phi^{(t)}-J_{\phi(t)}(x_m)\Delta\theta^{(t)}||^2,
\end{align}
but this would dramatically increase computational complexity and is practically infeasible.
Interestingly, our experiments reveal that no such explicit constraint is necessary: the standard embedding loss
\begin{align}
    L_{emb}=||sg(E(x_i))-c_{q_i}||^2
\end{align}
alone is sufficient to guide $P_\phi (\cdot)$ so that nearly all codebook entries remain active.

Across all tested variants of $P(\cdot)$, this implicit alignment consistently improves both codebook utilization and reconstruction performance.
\subsubsection{Design of $P_\phi(\cdot)$}
The mapping $P_\phi(\cdot)$ must be sufficiently generalizable to produce a transformed codebook
\begin{align}
    C^\prime=P_\phi(C)\in \mathbb{R}^d
\end{align}
without introducing structural constraints that would break the convergence conditions of VQ to k-means (as proven in Supplementary 8
). Notably, this is the key different between TransVQ and methods from~\cite{RN13, RN14}. They force $C^\prime=WC$, which adds another condition, making VQ not converge to the k-means solution anymore. A simple example is that we set the dimension of $C$ as 1, then $C^\prime=wC$. Since $C$ is frozen, $C\prime$ is unlikely to be the k-means solution of $E(x)$.
For this reason, we adopt a lightweight transformer block as $P_\phi (\cdot)$ :
\begin{itemize}
    \item Each code vector is treated as a token,
    \item Passed through a single-head linear attention layer,
    \item Followed by a small MLP layer.
\end{itemize}
The output defines the transformed codebook $C^\prime$.

\section{Experiments}
Although our approach can be extended to classification and generative tasks involving VQ, in this work we focus on evaluating NS-VQ and TransVQ within the VQ-VAE framework for image reconstruction. Since our goal is to validate the proposed hypotheses, extensive training under varied hyperparameter settings is required. This makes it impractical to conduct experiments on very large-scale datasets such as ImageNet. Instead, we adopt the CelebA-HQ (256×256) dataset, which contains 30,000 high-quality celebrity face images resampled to 256 pixels. Originally introduced by NVIDIA \cite{RN17}, this dataset is widely used for generative modeling and contains only images without attributes or labels. Additional results on ImageNet are provided in the Supplementary 12
.
The encoder and decoder architectures of VQ-VAE follow the design shown in Tab~\ref{tab:1 encoder_decoder}. We employ MSE loss as the reconstruction objective, and evaluate reconstruction quality using rFID, LPIPS, and SSIM metrics. We set batch size as 16, learning rate as 0.0005, use Adam optimizer.
For NS-VQ, the hyperparameter $2\sigma^2$ was initialized to 1 and decayed by a factor of 0.9 after each epoch. For TransVQ, we used a single linear transformer layer with one attention head, a model dimension of 256, and an MLP ratio of 2. All models were trained for 40 epochs on the training set, and their performance was evaluated on the validation set.
\begin{table}[t]
  \caption{Encoder and decoder architecture of the proposed model.}
  \label{tab:1 encoder_decoder}
  \centering
  \footnotesize  
  \setlength{\tabcolsep}{3pt}  
  \renewcommand{\arraystretch}{1.1}  
  \begin{tabular}{@{}p{0.46\columnwidth}p{0.46\columnwidth}@{}}
    \toprule
    \textbf{Encoder} & \textbf{Decoder} \\
    \midrule
    Conv2d (3$\!\rightarrow\!$128, $k$=4, $s$=2, $p$=1), LeakyReLU &
    Conv2d (64$\!\rightarrow\!$256, $k$=3, $s$=1, $p$=1), LeakyReLU \\

    Conv2d (128$\!\rightarrow\!$256, $k$=4, $s$=2, $p$=1), LeakyReLU &
    ResidualLayer $\!\times\!$6 (256$\!\rightarrow\!$256) \\

    Conv2d (256$\!\rightarrow\!$256, $k$=3, $s$=1, $p$=1), LeakyReLU &
    LeakyReLU \\

    ResidualLayer $\!\times\!$6 (256$\!\rightarrow\!$256) &
    ConvTranspose2d (256$\!\rightarrow\!$128, $k$=4, $s$=2, $p$=1), LeakyReLU \\

    LeakyReLU &
    ConvTranspose2d (128$\!\rightarrow\!$3, $k$=4, $s$=2, $p$=1), Tanh \\

    Conv2d (256$\!\rightarrow\!$64, $k$=1, $s$=1), LeakyReLU & -- \\
    \bottomrule
  \end{tabular}
\end{table}

\section{Results}
\begin{figure}[t]
  \centering
  \begin{subfigure}{0.8\linewidth}
    \centering
    \includegraphics[width=\linewidth]{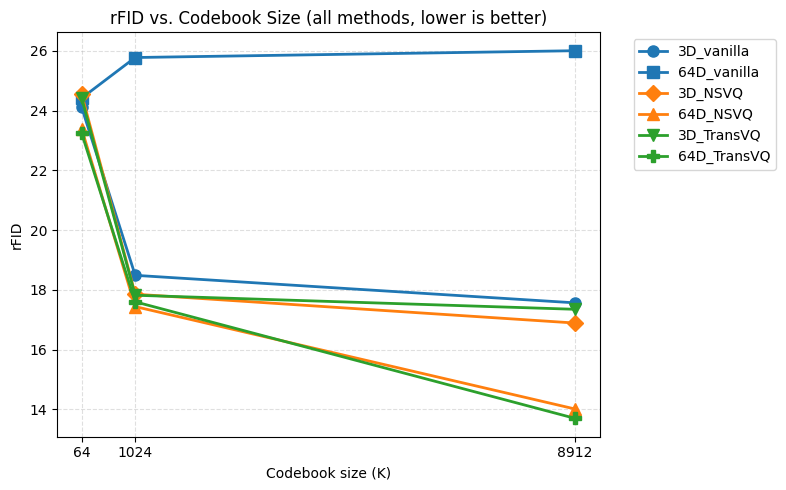}  
    \caption{rFID vs. Codebook Size (lower is better).}
    \label{fig:rfid}
  \end{subfigure}
  \hfill
  \begin{subfigure}{0.8\linewidth}
    \centering
    \includegraphics[width=\linewidth]{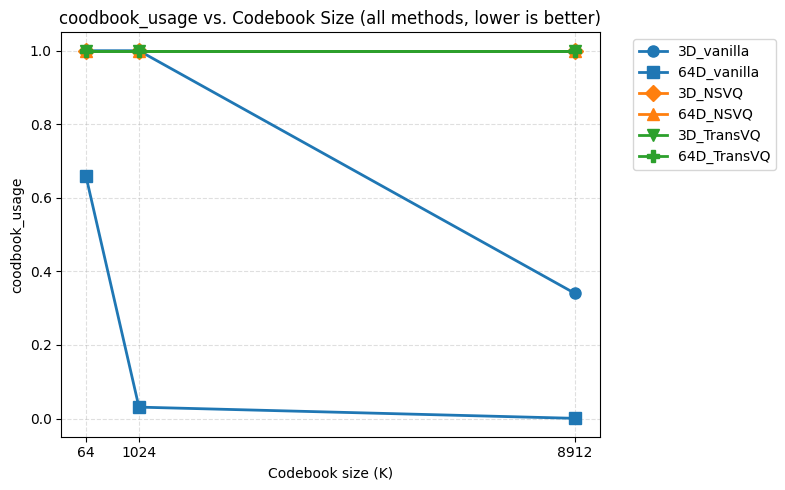}  
    \caption{Codebook Usage vs. Codebook Size (higher is better).}
    \label{fig:usage}
  \end{subfigure}
  \caption{
  Comparison of the proposed NS-VQVAE and TransVQVAE with VQGAN-FC~\cite{RN3} under varying codebook sizes.  
  (a) rFID comparison showing both NS-VQVAE and TransVQVAE consistently reduce reconstruction error compared with VQGAN-FC~\cite{RN3}.  
  (b) Codebook utilization indicating that both proposed methods maintain nearly full codebook usage, effectively preventing codebook collapse.
  }
  \label{fig:rfid_usage}
\end{figure}

\begin{figure}[t]
  \centering
  \includegraphics[width=0.85\linewidth]{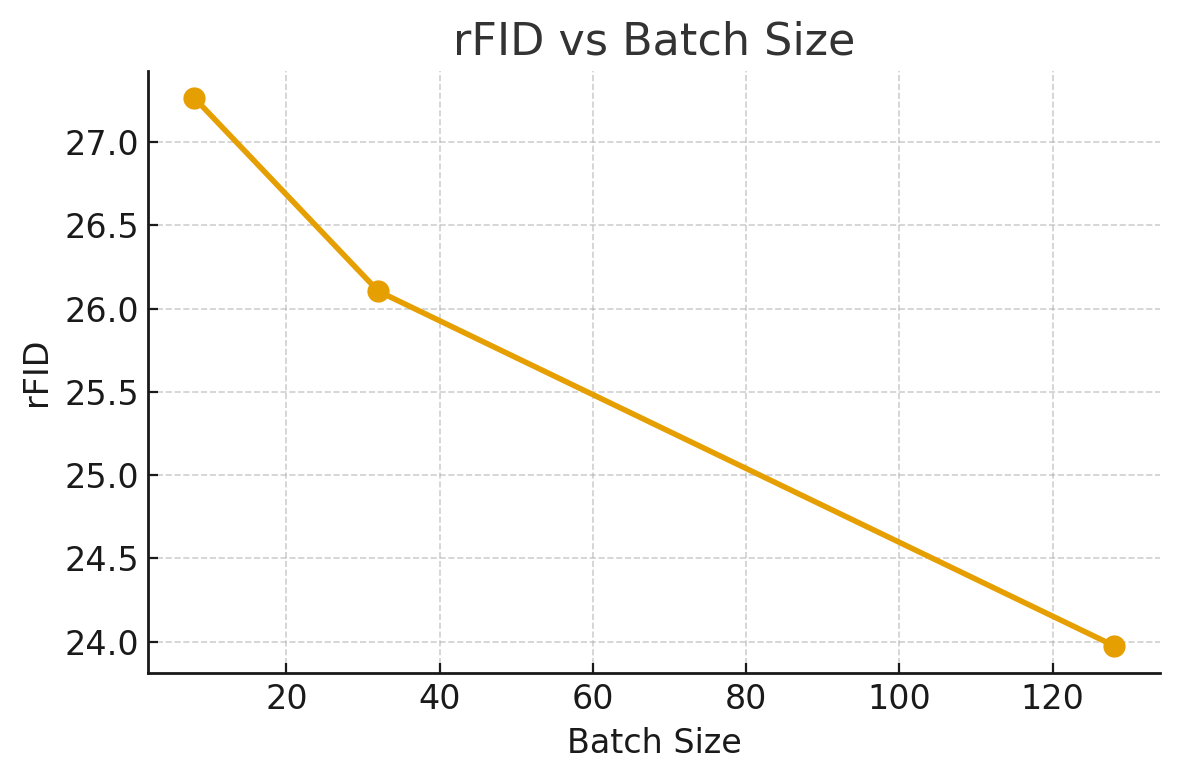}  
  \caption{
  rFID curves of standard VQ-VAE with codebook size and code dimension fixed at 64, evaluated under different batch sizes. 
  Larger batch sizes lead to lower rFID values, consistent with our theoretical analysis that larger batches provide more stable codebook updates.
  }
  \label{fig:rfid_batchsize}
\end{figure}

From Fig.~\ref{fig:rfid_usage}, we observe that NS-VQ and TransVQ do not suffer from the counterintuitive phenomenon seen in VQGAN-FC~\cite{RN3}, where reducing the codebook dimension improves reconstruction performance. Instead, our methods increase codebook utilization as predicted by theory, leading to improved reconstruction quality. Tabs.~\ref{tab:vq_3d} and ~\ref{tab:vq_64d} provides a detailed comparison of rFID across different codebook sizes and embedding dimensions, along with the corresponding codebook utilization. Tab.~\ref{tab:vq_detailed} further compares rFID, LPIPS, and SSIM for different VQ methods under their optimal codebook configurations. Finally, Fig.~\ref{fig:rfid_batchsize} shows the rFID curves of the standard VQ-VAE, with the codebook size and dimension fixed at 64, under varying batch sizes. The consistent decrease in rFID with larger batch sizes empirically supports our theoretical analysis.

\begin{table}[t]
  \caption{Comparison of different VQ methods with 3D code dimension under varying codebook sizes. Results are reported in terms of reconstruction FID (rFID, lower is better) and codebook usage (higher is better).}
  \label{tab:vq_3d}
  \centering
  \footnotesize
  \setlength{\tabcolsep}{4pt}
  \renewcommand{\arraystretch}{1.1}
  \begin{tabular}{@{}lcccccc@{}}
    \toprule
    \textbf{Method} & 
    \multicolumn{2}{c}{\textbf{64}} &
    \multicolumn{2}{c}{\textbf{1024}} &
    \multicolumn{2}{c}{\textbf{8912}} \\
    \cmidrule(lr){2-3} \cmidrule(lr){4-5} \cmidrule(lr){6-7}
     & rFID & Usage & rFID & Usage & rFID & Usage \\
    \midrule
    VQGAN-FC~\cite{RN3} & 24.10 & 1.00 & 18.48 & 1.00 & 17.57 & 0.34 \\
    VQVAE2~\cite{razavi2019generating} & 53.25 & 1.00 & 31.17 & 0.40 & 24.24 & 0.60 \\
    VQGAN-EMA~\cite{esser2021taming} & 60.86 & 1.00 & 17.86 & 1.00 & 17.93 & 1.00 \\
    SimVQ~\cite{esser2021taming} & \textbf{23.39} & 1.00 & 23.35 & 1.00 & 21.69 & 1.00 \\
    \textbf{NS-VQ} & 24.55 & 1.00 & 17.86 & 1.00 & \textbf{16.89} & 1.00 \\
    \textbf{TransVQ} & 24.42 & 1.00 & \textbf{17.82} & 1.00 & 17.35 & 1.00 \\
    \bottomrule
  \end{tabular}
\end{table}

\begin{table}[t]
  \caption{Comparison of different VQ methods with 64D code dimension under varying codebook sizes. Results are reported in terms of reconstruction FID (rFID, lower is better) and codebook usage (higher is better).}
  \label{tab:vq_64d}
  \centering
  \footnotesize
  \setlength{\tabcolsep}{4pt}
  \renewcommand{\arraystretch}{1.1}
  \begin{tabular}{@{}lcccccc@{}}
    \toprule
    \textbf{Method} & 
    \multicolumn{2}{c}{\textbf{64}} &
    \multicolumn{2}{c}{\textbf{1024}} &
    \multicolumn{2}{c}{\textbf{8912}} \\
    \cmidrule(lr){2-3} \cmidrule(lr){4-5} \cmidrule(lr){6-7}
     & rFID & Usage & rFID & Usage & rFID & Usage \\
    \midrule
    VQGAN-FC~\cite{RN3} & 24.41 & 0.66 & 25.77 & 0.03 & 26.00 & 0.00 \\
    VQVAE2~\cite{razavi2019generating} & 85.86 & 0.82 & 36.91 & 0.57 & 25.87 & 0.28 \\
    VQGAN-EMA~\cite{esser2021taming} & 159.21 & 1.00 & 20.81 & 1.00 & 19.39 & 1.00 \\
    SimVQ~\cite{RN13} & 23.40 & 1.00 & 17.77 & 1.00 & 14.37 & 1.00 \\
    \textbf{NS-VQ} & 23.39 & 1.00 & \textbf{17.44} & 1.00 & 14.01 & 1.00 \\
    \textbf{TransVQ} & \textbf{23.24} & 1.00 & 17.60 & 1.00 & \textbf{13.70} & 1.00 \\
    \bottomrule
  \end{tabular}
\end{table}

\begin{table}[t]
  \caption{Detailed comparison of VQ methods at their best-performing configurations (as determined by rFID in Table~\ref{tab:vq_3d} and Table~\ref{tab:vq_64d}). 
  Results are reported in terms of SSIM (higher is better), LPIPS (lower is better), and rFID (lower is better).}
  \label{tab:vq_detailed}
  \centering
  \footnotesize
  \setlength{\tabcolsep}{5pt}
  \renewcommand{\arraystretch}{1.1}
  \begin{tabular}{@{}lccccc@{}}
    \toprule
    \textbf{Method} & \textbf{Codebook Size} & \textbf{Code Dim} & \textbf{SSIM} & \textbf{LPIPS} & \textbf{rFID} \\
    \midrule
    VQGAN-FC~\cite{RN3} & 8912 & 3  & 0.93 & 0.022 & 17.57 \\
    VQVAE2~\cite{razavi2019generating} (3D) & 8912 & 3  & 0.89 & 0.042 & 24.24 \\
    VQGAN-EMA~\cite{esser2021taming} & 1024 & 3  & 0.92 & 0.024 & 17.86 \\
    SimVQ~\cite{RN13} & 8912 & 64 & 0.93 & 0.017 & 14.37 \\
    \textbf{NS-VQ} & 8912 & 64 & \textbf{0.96} & \textbf{0.015} & 14.01 \\
    \textbf{TransVQ} & 8912 & 64 & 0.94 & \textbf{0.015} & \textbf{13.70} \\
    \bottomrule
  \end{tabular}
\end{table}

\section{Discussion}
In this work, we restrict our comparison to VQ methods that converge to the k-means solution. Many recent variants have deviated from the stochastic approximation of k-means; for example, FSQ~\cite{RN16} performs direct quantization of latent features rather than clustering, while LFQ~\cite{RN7} can be regarded as sign or directional clustering. These methods are therefore excluded from our evaluation. In Eq. 17, we proposed replacing $\Delta E(x_i )$ with $(x_i-c_{q_j})$. In the Supplementary 10 
provides a derivation for directly using $\Delta E(x_i)$, which also improves traditional VQ in practice and may be of interest for further exploration.

We also note certain limitations. NS-VQ, while theoretically grounded, requires tuning the hyperparameter $2\sigma^2$, which typically must decay with training iterations—making the optimization process more complex. In contrast, TransVQ is easier to tune but introduces additional computational cost. Interestingly, we observed that replacing the transformer block with a two-layer MLP slightly reduces performance but significantly improves speed, which may be useful in practice.

Finally, we would like to point out an open question regarding the common practice of VQ initialization and training. Most existing works—including ours—first initialize the codebook using k-means and then perform online updates through the VQ layer until convergence to the k-means solution in the latent space. However, this raises a natural question: if a global k-means step is already performed, why not simply train a conventional autoencoder, freeze the encoder after training, and then apply k-means on the latent features to obtain the codebook? This approach would even reduce the overall computational cost.
One possible explanation is that the quantization layer in VQ-VAE inherently prevents the autoencoder from degenerating into an identity mapping, a problem often encountered in standard AEs. Nevertheless, to the best of our knowledge, there has been no theoretical or empirical study that explicitly analyzes this effect. Understanding why the VQ layer improves representation learning over post-hoc k-means clustering represents an important future direction for VQ-VAE research.

\section{Conclusion}
In this work, we revisited vector quantization from a theoretical perspective and identified the non-stationary nature of VQ-VAE as the fundamental cause of codebook collapse. Building on this insight, we proposed two new approaches, NS-VQ and TransVQ, that propagate encoder drift to non-selected codes or adapt the entire codebook through a learnable mapping. Both methods preserve the convergence conditions of VQ while significantly improving codebook utilization and reconstruction quality.

Extensive experiments on CelebA-HQ and additional validations demonstrate that NS-VQ and TransVQ consistently outperform standard VQ variants in terms of rFID, LPIPS, and SSIM, while maintaining near-complete codebook usage. Beyond improving empirical performance, our analysis provides a stronger theoretical foundation for understanding and addressing codebook collapse.

Looking forward, the proposed methods can generalize beyond image reconstruction to a wide range of generative and multimodal tasks where VQ serves as the discrete interface, including large-scale visual-language models. Our work thus bridges the gap between theory and practice in vector quantization, offering new directions for building more scalable and reliable VQ-based models.

Although our methods enhance stability and utilization, they introduce additional hyperparameters (e.g., $2\sigma^2$ and transformer depth) that require careful tuning. Future work could focus on developing adaptive or self-regularizing mechanisms to automatically control these parameters during training. Moreover, integrating NS-VQ and TransVQ into diffusion, autoregressive, and multimodal architectures may further reveal their potential for large-scale generative modeling. Another promising direction is to explore dynamic or hierarchical codebook expansion, allowing the model to flexibly adjust its representational capacity as training progresses. Lastly, a deeper theoretical study of how quantization interacts with gradient flow and representation learning could provide a unified framework for next-generation VQ-based models.

{
    \small
    \bibliographystyle{IEEEtran}
    \bibliography{main}
}
\clearpage
\setcounter{page}{1}
\maketitlesupplementary

\section{Proof that VQ Converges to the \textit{k}-Means Solution}
\label{sec:proof_vq_kmeans}

Without loss of generality, assume data $x \in \mathbb{R}^d$ is sampled i.i.d. from a distribution $P$.  
The same argument applies to the finite empirical case, replacing integrals by finite sums.  
We maintain $K$ centroids (or codes) $C = (c_1, \dots, c_K) \in (\mathbb{R}^d)^K$.  
Given $C$, define the Voronoi partition:
\[
V_k(C) = \{x : \|x - c_k\| \le \|x - c_j\|, \; \forall j \}.
\]
The expected distortion (quantization error) is
\begin{align}
J(C)
&= \mathbb{E}_{x \sim P} \!\left[ \min_k \|x - c_k\|^2 \right]  \notag \\
&= \sum_{k=1}^K \int_{V_k(C)} \|x - c_k\|^2 \, dP(x).
\tag{A1}
\end{align}

The \textit{k}-means problem (and equivalently, VQ codebook design) seeks to minimize $J(C)$.  
Away from measure-zero tie boundaries, the partition does not change under small centroid moves, so we can differentiate $J$ at regular points:
\begin{align}
\frac{\partial J}{\partial c_k}(C)
&= \frac{\partial}{\partial c_k}
  \sum_{k=1}^K \int_{V_k(C)} \|x - c_k\|^2 \, dP(x)  \notag \\
&= -2 \int_{V_k(C)} (x - c_k) \, dP(x).
\tag{A2}
\end{align}
Setting the gradient to zero gives
\begin{align}
\int_{V_k(C)} x \, dP(x)
&= c_k \int_{V_k(C)} dP(x), \notag \\
&\Longleftrightarrow \;
c_k = \mathbb{E}[x \mid x \in V_k(C)].
\tag{A3}
\end{align}
Thus, the stationary points of \textit{k}-means are those $C^*$ for which every centroid equals the conditional mean of its Voronoi cell—the usual M-step in \textit{k}-means.

\paragraph{Stochastic Online Update.}
Consider the standard “winner-take-all” online VQ update for a sample $x_t \sim P$:
\begin{align}
k_t(x_t) &= \arg\min_j \|x_t - c_j^{(t)}\|, \tag{A4}\\
c_k^{(t+1)} &= c_k^{(t)} + \eta_t \, \mathbf{1}\{k = k_t(x_t)\} (x_t - c_k^{(t)}),
\tag{A5}
\end{align}
where $\eta_t$ is the step size.  
Taking conditional expectation given $C^{(t)}$:
\begin{equation}
\mathbb{E}[c_k^{(t+1)} - c_k^{(t)} \mid C^{(t)}]
  = \eta_t \, h_k(C^{(t)}),
\tag{A6}
\end{equation}
where
\begin{align}
h_k(C)
&\triangleq \mathbb{E}\!\left[\mathbf{1}\{x \in V_k(C)\}(x - c_k)\right] \notag \\
&= \int_{V_k(C)} (x - c_k) \, dP(x).
\tag{A7}
\end{align}
Comparing Eq.~(A7) with Eq.~(A2), we have
\[
\frac{\partial J}{\partial c_k}(C) = -2 h_k(C).
\]
Hence, the mean update direction equals $-\tfrac{1}{2} \nabla J(C)$, meaning online VQ is a stochastic approximation (SA) to gradient descent of the \textit{k}-means objective.

We can rewrite the recursion in SA form:
\begin{equation}
c_k^{(t+1)} = c_k^{(t)} + \eta_t \big(h_k(C^{(t)}) + \xi_{t,k}\big),
\tag{A8}
\end{equation}
with a martingale-difference noise term
\begin{align}
\xi_{t,k}
&= \mathbf{1}\{x_t \in V_k(C^{(t)})\}(x_t - c_k^{(t)}) \notag \\
&\quad - h_k(C^{(t)}),
\quad \mathbb{E}[\xi_{t,k} \mid C^{(t)}] = 0.
\tag{A9}
\end{align}

\paragraph{Convergence Conditions.}
The following assumptions hold:
\begin{itemize}
\item \textbf{Step sizes (Robbins–Monro):}
  $\sum_t \eta_t = \infty$, \;
  $\sum_t \eta_t^2 < \infty$.
\item \textbf{Boundedness:}
  Data have bounded second moment and centroids are projected to a compact set.
\item \textbf{Regularity:}
  $h(C)$ is locally Lipschitz away from tie boundaries, which form a measure-zero set.
\end{itemize}

Under these conditions, by standard SA/ODE theory~\cite{RN18,RN19},  
the discrete iterates track the flow of the ODE
\begin{equation}
\dot{C}(t) = h(C(t)) = -\tfrac{1}{2} \nabla J(C(t)).
\tag{A10}
\end{equation}
Using $J$ as a Lyapunov function:
\begin{align}
\frac{d}{dt} J(C(t))
&= \langle \nabla J(C(t)), \dot{C}(t) \rangle \notag \\
&= -\tfrac{1}{2} \|\nabla J(C(t))\|^2 \le 0.
\tag{A11}
\end{align}
Therefore, trajectories converge to the stationary set $\{\nabla J = 0\}$.  
From Eq.~(A2), any limit point $C^*$ satisfies
\[
c_k^* = \mathbb{E}[x \mid x \in V_k(C^*)], \quad \forall k,
\]
which is a \textit{k}-means fixed point (typically a local minimum).  
Hence, online VQ with Robbins–Monro step sizes converges almost surely to a \textit{k}-means solution (local optimum).
\section{Modified Straight-Through Estimator (STE)}
\label{sec:modified_ste}

In the main text, we revisited the update rule proposed in~\cite{RN9} [Eq.~(21)],
which introduced a ``delay-free'' correction term for updating
$c_{q_i}^{(0)}$ based on $E(x_i)^{(1)} - c_{q_i}^{(0)}$.
That method, however, required tuning an additional hyperparameter $v$,
which limits its practical applicability.
Here, we derive a similar correction from first principles
without introducing new hyperparameters.

\subsection{Revisiting the Embedding Loss}
In VQ-VAE, the embedding loss at iteration $t$ is
\begin{equation}
L_{\text{emb}}^{(t)}(x_n, c_{q_n})
= \frac{1}{d} \left\| \operatorname{sg}\!\big(E(x_n)^{(t)}\big)
  - c_{q_n}^{(t)} \right\|^2,
\tag{A16}
\end{equation}
which encourages $c_{q_n}^{(t)}$ to move toward $E(x_n)^{(t)}$.
After one iteration, the encoder updates according to
\begin{equation}
E(x_n)^{(t+1)} = E(x_n)^{(t)} +
\lambda \frac{\partial L}{\partial E(x_n)^{(t)}}.
\tag{A17}
\end{equation}
The natural question arises:  
why not directly update $c_{q_n}^{(t)}$ so that
it also moves toward the newly updated encoder output $E(x_n)^{(t+1)}$.

\subsection{Delay-Free Correction Term}
In that case, we define
\begin{equation}
L_{\text{emb}}^{(t)'}(x_n, c_{q_n})
= \frac{1}{d}
  \left\| \operatorname{sg}\!\big(E(x_n)^{(t+1)}\big)
  - c_{q_n}^{(t)} \right\|^2.
\tag{A18}
\end{equation}
The gradient with respect to $c_{q_n}^{(t)}$ becomes
\begin{align}
\frac{\partial L_{\text{emb}}^{(t)'}}{\partial c_{q_n}^{(t)}}
&= \frac{2}{d}\left(E(x_n)^{(t+1)} - c_{q_n}^{(t)}\right) \notag\\
&= \frac{2}{d}\left(E(x_n)^{(t)} - c_{q_n}^{(t)}\right)
 + \frac{2\lambda}{d} \frac{\partial L}{\partial E(x_n)^{(t)}}.
\tag{A19}
\end{align}
The first term can be obtained directly from
the standard embedding loss $L_{\text{emb}}^{(t)}$,
while the second term can be propagated to the codebook through
the straight-through estimator (STE).

\subsection{Modified Quantizer with STE Injection}
Using the STE formulation, the quantized output is
\begin{equation}
Q_e = E(x_n)^{(t)} +
\operatorname{sg}\!\big(c_{q_n}^{(t)} - E(x_n)^{(t)}\big).
\tag{A20}
\end{equation}
To inject the additional gradient flow from $E(x_n)^{(t)}$
into $c_{q_n}^{(t)}$, we modify the quantizer as
\begin{align}
Q_e &= E(x_n)^{(t)} +
\operatorname{sg}\!\big(c_{q_n}^{(t)} - E(x_n)^{(t)}\big) \notag\\
&\quad + \tfrac{2}{d} c_{q_n}^{(t)} -
  \operatorname{sg}\!\big(\tfrac{2}{d} c_{q_n}^{(t)}\big).
\tag{A21}
\end{align}
Here, $Q_e$ denotes the quantization output of the VQ layer.
This modification ensures that the gradient of
$E(x_n)^{(t)}$ automatically propagates into $c_{q_n}^{(t)}$,
achieving a delay-free correction without introducing any new hyperparameters.

\section{Alternative Update Using $\Delta E(x)$}
\label{sec:alternative_update}

In the main text, we proposed a practical approximation that replaces
$\Delta E(x_i)$ with $(x_i - c_{q_j})$.
Here, we provide the complementary formulation that directly propagates $\Delta E(x)$
to the codebook updates.

\subsection{Recap of the Approximation}
From the NTK-based derivation, we have
\begin{equation}
\Delta E(x) \approx
\exp\!\left(-\frac{\|x - x_i\|^2}{2\sigma^2}\right)
\, \Delta E(x_i).
\tag{A12}
\end{equation}
Since the future sample $x$ is unavailable at the current iteration,
we approximate the distance using code vectors:
\begin{equation}
\|x - x_i\|^2 \approx \|c_q - x_i\|^2.
\tag{A13}
\end{equation}
This approximation allows all codebook entries to be updated in parallel
without waiting for additional samples.

\subsection{STE-Based Injection}
Interestingly, we can inject $\Delta E(x_i)$ into the
straight-through estimator (STE) via the stop-gradient trick.
Specifically, the quantized embedding can be expressed as
\begin{align}
Q_e
&= E(x_i)
 + \operatorname{sg}\!\big(c_{q_i} - E(x_i)\big) \notag\\
&\quad + \sum_{q_n \ne q_i}
   \big(w_n c_{q_n} - \operatorname{sg}(w_n c_{q_n})\big),
\tag{A14}
\end{align}
where the weights are defined as
\begin{equation}
w_n = \exp\!\left(-\frac{\|c_{q_n} - x_i\|^2}{2\sigma^2}\right).
\tag{A15}
\end{equation}

This formulation distributes a fraction of $\Delta E(x_i)$
across all non-selected codes,
thereby alleviating the dead-code problem
by ensuring that each code receives gradient information
proportional to its similarity to the current sample.

\section{Toy Demos for Non-Stationary Vector Quantization}
\label{sec:toy_demos}

This appendix provides three toy experiments that illustrate how non-stationarity
affects vector quantization and how our proposed NS-VQ stabilizes training
under dynamic data distributions.

\subsection{Toy Demo 1 — Translation Motion}
\label{sec:toy_translation}

\paragraph{Purpose.}
Compare two codebook-update rules for vector quantization on a simple
non-stationary 2D dataset where a global drift variable $ {\theta}$ evolves during training:
\begin{itemize}
    \item EMA (vanilla VQ update method)
    \item NS-VQ (our proposed method)
\end{itemize}

\paragraph{Data and Non-Stationarity.}
\begin{align*}
& X_{\text{init}} \sim \mathcal{N}(0, I_2), \quad N = 1500, \quad \text{noise scale}=1.0, \\
&  {\theta} \in \mathbb{R}^2, \quad  {\theta}_0 = [0, 0].
\end{align*}
For each mini-batch with indices \texttt{idxs}:
\[
X_b = X[\texttt{idxs}] + {\theta}, \qquad
Y_t^{(b)} = X_{\text{init}}[\texttt{idxs}] + (10, 10).
\]
The drift variable is updated by
\[
 {\theta} \leftarrow  {\theta} + r \cdot \mathrm{mean}(Y_t^{(b)} - X_b),
\quad r = 0.1.
\]

\paragraph{Codebook and Assignment.}
\[
C \in \mathbb{R}^{16 \times 2}, \quad C \sim \mathcal{N}(0, 1), \quad
w_i = \arg\min_k \|x - c_k\|_2^2.
\]

\paragraph{Update Rules.}

\textbf{(1) EMA Update.}
If cluster $k$ has assigned samples in the current batch:
\begin{align*}
\bar{y}_k &= \frac{1}{|\{i : w_i = k\}|} \sum_{i:w_i=k} X_b^{(i)}, \\
C_k &\leftarrow (1-\alpha) C_k + \alpha \bar{y}_k,
\quad \alpha = 0.3.
\end{align*}

\textbf{(2) NS-VQ Online Update.}
Processed sample-by-sample within each mini-batch using temperature $\tau$
and learning rate $\text{lr}$:
\begin{align*}
d_k &= \|C_k - C_w\|^2, \qquad
\omega_k = \frac{e^{-d_k / \tau}}{\sum_j e^{-d_j / \tau}}, \\
C_w &\leftarrow C_w + \text{lr}(x - C_w), \\
C_k &\leftarrow C_k + \text{lr}\,\omega_k (x - C_k), \quad k \ne w.
\end{align*}
After each epoch:
\[
\tau \leftarrow \tfrac{1}{2}\tau, \qquad \text{lr} \leftarrow 0.9\,\text{lr}.
\]


\subsection{Toy Demo 2 — Expansion / Scaling}
\label{sec:toy_expansion}

\paragraph{Purpose.}
Compare two codebook-update rules on a non-stationary 2D dataset
where the data undergo a global linear scaling learned online via a matrix $A$.

\paragraph{Data and Target Mapping (Fixed).}
\[
X_{\text{init}} \sim \mathcal{N}(0, I_2), \quad N = 1500.
\]
Draw $M \in \mathbb{R}^{2\times2}$ such that $\|M\|_2 > 1$ and define the fixed targets:
\[
Y_t = X_{\text{init}} M^\top.
\]

\paragraph{Learned State and Batch Variables.}
Initialize $A = I$.  
For a mini-batch with indices \texttt{idxs}:
\begin{align*}
X_b &= X[\texttt{idxs}], \quad Y_b = Y_t[\texttt{idxs}], \\
A X_b &= X_b A^\top, \quad E = Y_b - A X_b.
\end{align*}
Update of $A$ (exact code):
\begin{align*}
g_A &= \frac{2}{B} E^\top X_b, \quad \text{(descent direction of $-\nabla_A \|Y_b - A X_b\|_F^2$)} \\
A &\leftarrow A + r g_A, \quad r = 0.1.
\end{align*}
All other settings are identical to Appendix~\ref{sec:toy_translation}.
We also tested a \emph{shrink} variant by setting $\|M\|_2 < 1$.


\subsection{Toy Demo 3 — Split (Quadrant Separation)}
\label{sec:toy_split}

\paragraph{Purpose.}
Compare the two update rules on a 2D dataset where samples split by sign
into opposite quadrants and drift scales by coordinate sign.

\paragraph{Data, Target, and Non-Stationarity.}
\begin{align*}
X_{\text{init}} &\sim \mathcal{N}( {\mu}, \sigma^2 I_2),
\quad  {\mu} = [0.5,\,0.5], \quad \sigma = 1.0, \\
Y_t &= X_{\text{init}} + 10 \cdot \operatorname{sign}(X_{\text{init}}).
\end{align*}
Initialize drift $ {\theta} = [0, 0]$.
For each batch:
\begin{align*}
X_b &= X[\texttt{idxs}], \\
X_b &\leftarrow X_b + \operatorname{sign}(X_b) \odot  {\theta}, \\
 {\theta} &\leftarrow  {\theta} + r \cdot \mathrm{mean}(Y_t^{(b)} - X_b),
\quad r = 0.1.
\end{align*}
For visualization, the full cloud at each step is
\[
X_b + \operatorname{sign}(X_b) \odot  {\theta}.
\]

\section{Results on ImageNet} \label{sec:result_on_imagenet}

We employ a Taming-style convolutional encoder–decoder architecture without self-attention, consisting of three resolution stages. The VQ-VAE model is trained on ImageNet-1K at a resolution of $256\times256$ using a batch size of 12 and Adam optimizers with $\beta_1=0.5$ and $\beta_2=0.9$. The base learning rate is set to $4.5\times10^{-6}$. The reconstruction objective combines mean squared error (MSE) and LPIPS loss. All hyperparameter settings for NS-VQ and TransVQ follow those described in Section~4. Each experiment is trained for four epochs.

\begin{table}[h]
\centering
\caption{VQ-VAE architecture (Taming-style, attention-free). The encoder and decoder are symmetric with residual connections.}
\begin{tabular}{p{0.45\linewidth} p{0.45\linewidth}}
\toprule
\textbf{Encoder} & \textbf{Decoder} \\
\midrule
Conv2d (3$\rightarrow$128, k=4, s=2, p=1), LeakyReLU & Conv2d (3$\rightarrow$512, k=3, s=1, p=1), LeakyReLU \\
ResidualBlock $\times$2 (128$\rightarrow$128) & ResidualBlock $\times$2 (512$\rightarrow$512) \\
Conv2d (128$\rightarrow$256, k=4, s=2, p=1), LeakyReLU & ConvTranspose2d (512$\rightarrow$256, k=4, s=2, p=1), LeakyReLU \\
ResidualBlock $\times$2 (256$\rightarrow$256) & ResidualBlock $\times$2 (256$\rightarrow$256) \\
Conv2d (256$\rightarrow$512, k=4, s=2, p=1), LeakyReLU & ConvTranspose2d (256$\rightarrow$128, k=4, s=2, p=1), LeakyReLU \\
ResidualBlock $\times$2 (512$\rightarrow$512) & ResidualBlock $\times$2 (128$\rightarrow$128) \\
Conv2d (512$\rightarrow$3, k=3, s=1, p=1) $\rightarrow$ quant\_conv (1$\times$1, 3$\rightarrow$embed\_dim=3) & post\_quant\_conv (1$\times$1, embed\_dim=3$\rightarrow$3) \\
--- & Conv2d (128$\rightarrow$3, k=3, s=1, p=1), Tanh \\
\bottomrule
\end{tabular}
\end{table}

\begin{table}[h]
\centering
\caption{Quantitative comparison on ImageNet-1K at $256\times256$ resolution. Lower rFID and LPIPS indicate better reconstruction quality and perceptual fidelity.}
\begin{tabular}{lcc}
\toprule
\textbf{Method} & \textbf{rFID} & \textbf{LPIPS} \\
\midrule
VQGAN~\cite{RN6} & 1.060 & 0.113 \\
VQGAN-FC~\cite{RN3} & 0.600 & 0.087 \\
VQGAN-EMA~\cite{razavi2019generating} & 1.270 & 0.124 \\
SimVQ~\cite{RN13} & 0.551 & 0.088 \\
NS-VQ (ours) & 0.561 & 0.087 \\
TransVQ (ours) & \textbf{0.432} & \textbf{0.067} \\
\bottomrule
\end{tabular}
\end{table}

\section{Qualitative Reconstructions}
\label{sec:qual_recon}

\begin{figure*}[t]
  \centering
  \begin{subfigure}{0.7\linewidth}
    \centering
    \includegraphics[width=\linewidth]{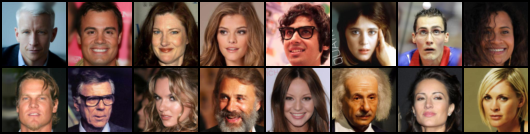}
    \caption{Ground truth (GT).}
    \label{fig:qual_gt}
  \end{subfigure}\vspace{4pt}

  \begin{subfigure}{0.7\linewidth}
    \centering
    \includegraphics[width=\linewidth]{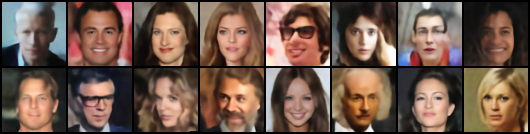}
    \caption{Vanilla VQ-VAE.}
    \label{fig:qual_vq}
  \end{subfigure}\vspace{4pt}

  \begin{subfigure}{0.7\linewidth}
    \centering
    \includegraphics[width=\linewidth]{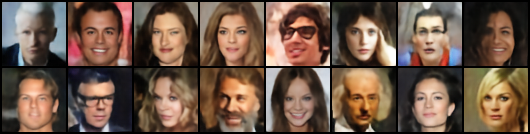}
    \caption{EMA-VQ.}
    \label{fig:qual_emavq}
  \end{subfigure}\vspace{4pt}

  \begin{subfigure}{0.7\linewidth}
    \centering
    \includegraphics[width=\linewidth]{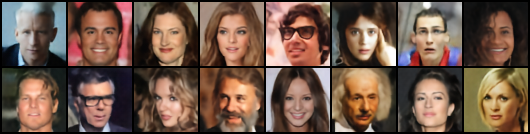}
    \caption{NS-VQ (ours).}
    \label{fig:qual_nsvq}
  \end{subfigure}\vspace{4pt}

  \begin{subfigure}{0.7\linewidth}
    \centering
    \includegraphics[width=\linewidth]{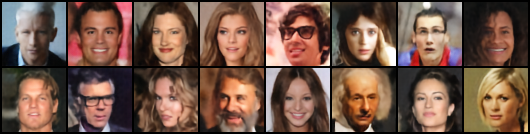}
    \caption{SimVQ.}
    \label{fig:qual_simvq}
  \end{subfigure}\vspace{4pt}

  \begin{subfigure}{0.7\linewidth}
    \centering
    \includegraphics[width=\linewidth]{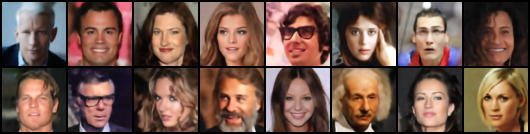}
    \caption{TransVQ (ours).}
    \label{fig:qual_transvq}
  \end{subfigure}

  \caption{
  \textbf{Qualitative reconstruction comparison.}
  Each strip shows the same set of identities. NS-VQ and TransVQ maintain sharp edges
  and facial micro-structure while reducing artifacts observed in vanilla and EMA-VQ.
  This aligns with the quantitative gains in rFID/LPIPS/SSIM reported in the main paper.
  }
  \label{fig:qual_recon}
\end{figure*}

Figure~\ref{fig:qual_recon} presents side-by-side reconstructions for six methods:
ground-truth (GT), vanilla VQ-VAE, EMA-VQ, NS-VQ (ours), SimVQ, and TransVQ (ours).
Consistent with the quantitative results in the main paper, both NS-VQ and TransVQ
recover finer facial detail and textures while avoiding over-smoothing. In particular,
NS-VQ reduces blocky artifacts by maintaining active code utilization, and TransVQ
preserves high-frequency cues (hair strands, eye contours) via its codebook transform.

\begin{landscape}
\begin{figure}[p]
  \centering
  \includegraphics[width=0.95\paperwidth]{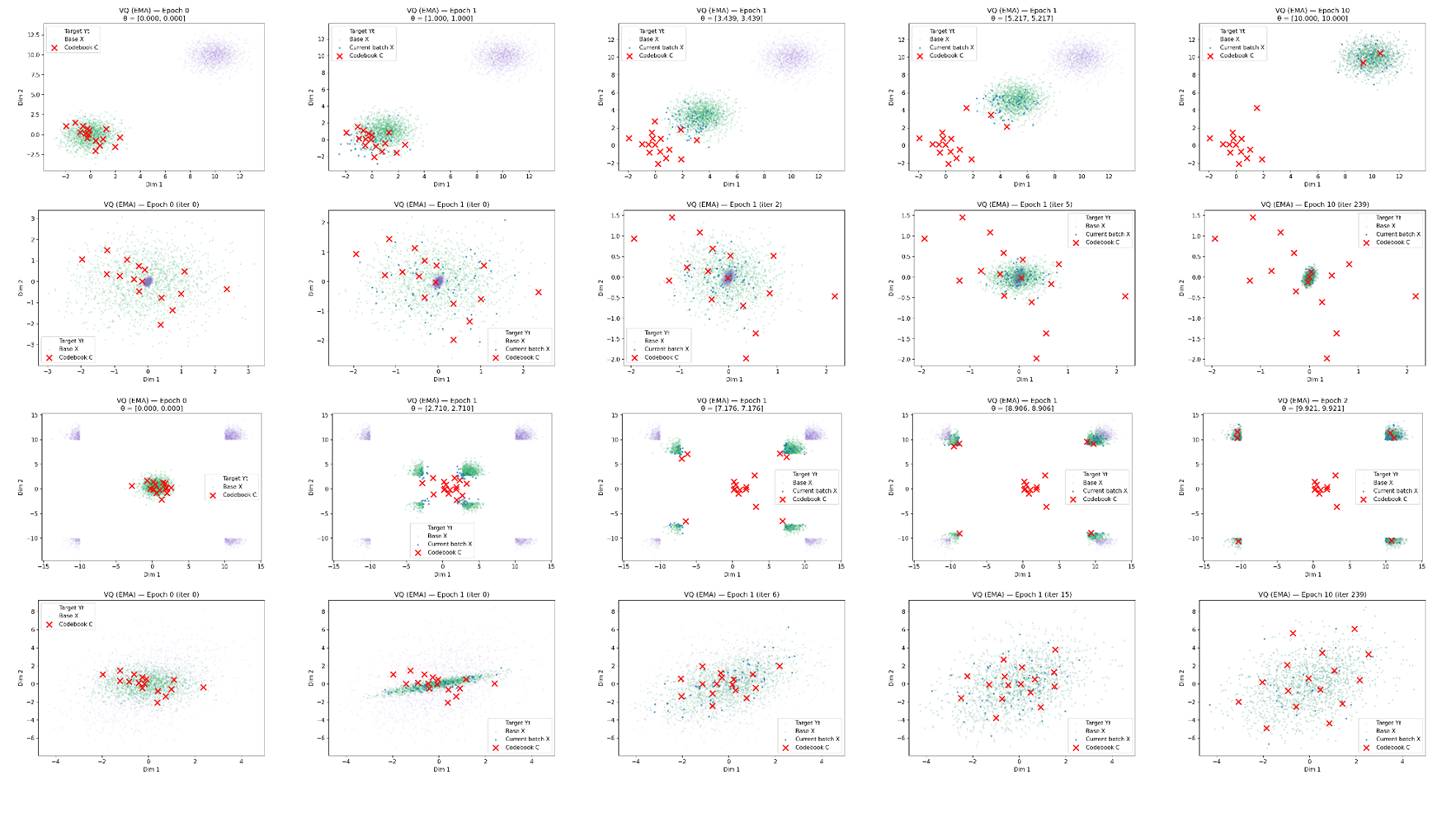} 
  \caption{
  \textbf{Toy Demo Visualization for Vanilla VQ.}
  Each row corresponds to one toy experiment:
  translation (1st row), shrink (2nd row), split (3rd row), and expansion (4th row).
  Each row contains five snapshots showing the convergence process across training steps.
  }
  \label{fig:toy_vq_vanilla}
\end{figure}
\end{landscape}

\begin{landscape}
\begin{figure}[p]
  \centering
  \includegraphics[width=0.95\paperwidth]{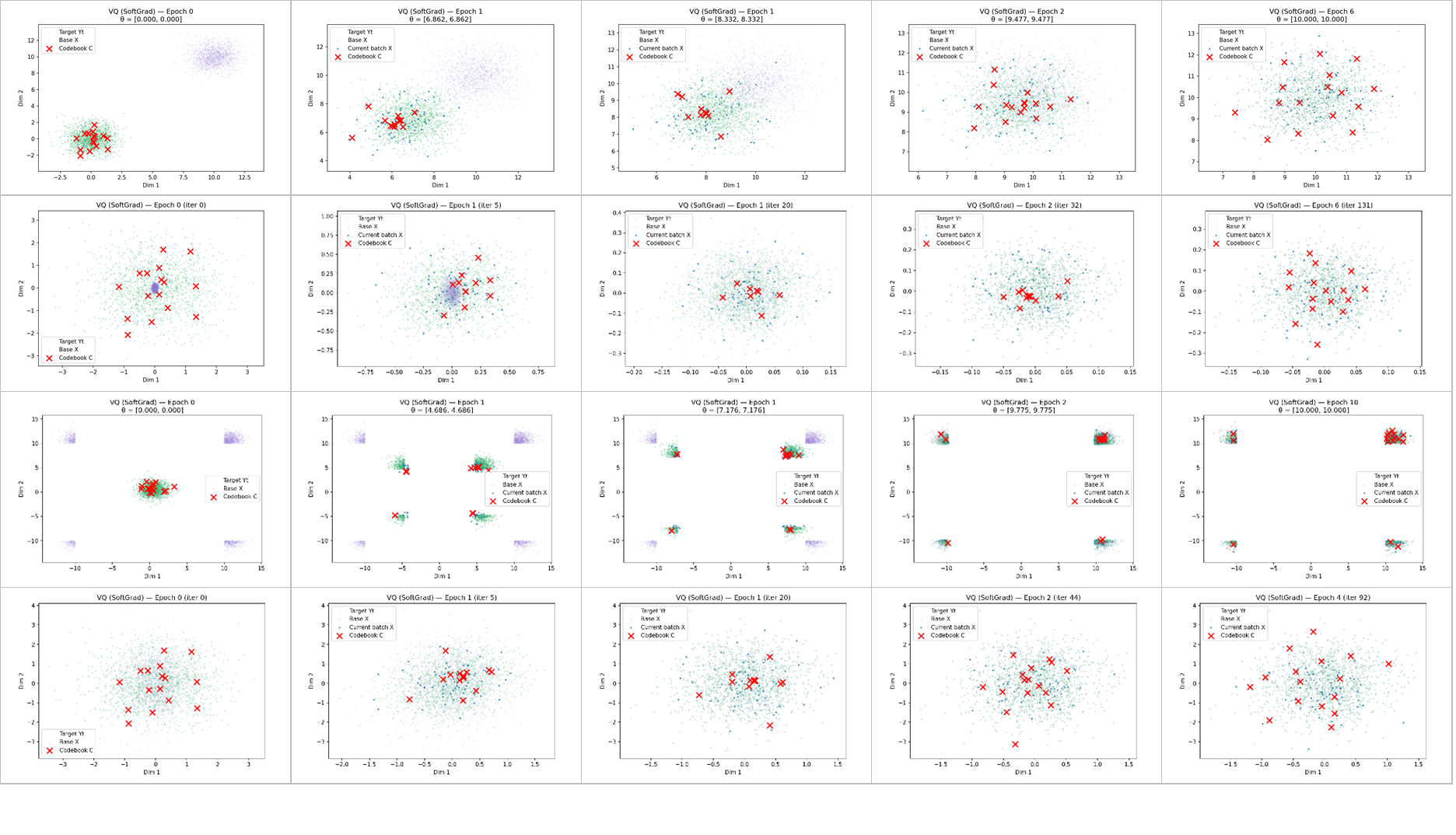} 
  \caption{
  \textbf{Toy Demo Visualization for NS-VQ.}
  Each row corresponds to one toy experiment:
  translation (1st row), shrink (2nd row), split (3rd row), and expansion (4th row).
  Each row contains five snapshots showing how NS-VQ maintains stability and codebook utilization during convergence.
  }
  \label{fig:toy_vq_nsvq}
\end{figure}
\end{landscape}


\end{document}